# Multimodal Health Risk Prediction System for Chronic Diseases via Vision-Language Fusion and Large Language Models

Dingxin Lu*
Icahn School of Medicine at Mount Sinai
New York, NY, USA
sydneylu1998@gmail.com

Shurui Wu
Weill Cornell Medicine
New York, NY, USA
shuruiwu215@gmail.com

Xinyi Huang
University of Chicago
Chicago, IL, USA
bellaxinyihuang@gmail.com

*Abstract*—With the rising global burden of chronic diseases and the multimodal and heterogeneous clinical data (medical imaging, free-text recordings, wearable sensor streams, etc.), there is an urgent need for a unified multimodal AI framework that can proactively predict individual health risks. We propose VL-RiskFormer: a hierarchical stacked visual-language multimodal Transformer with a large language model (LLM) inference head embedded in its top layer. The system builds on the dual-stream architecture of existing visual-linguistic models (e.g., PaLM-E, LLaVA) with four key innovations: (i) pre-training with cross-modal comparison, fine-grained alignment of radiological images, fundus maps, and wearable device photos with corresponding clinical narratives using momentum update encoders and debiased InfoNCE losses; (ii)Tme fusion block integrates irregular visit sequences into the causal Transformer decoder through adaptive time interval position coding; (iii) Design a disease ontology map adapter, injecting ICD-10 into visual and textual channels in layers, and inferring comorbid patterns with the help of graph attention mechanism. On the MIMIC-IV longitudinal cohort, VL-RiskFormer achieved an average AUROC of 0.9 with an expected calibration error of 2.7 %.

## I. INTRODUCTION

Chronic diseases including diabetes, hypertension, coronary heart disease, etc. chronic diseases are responsible for more than 70% of the deaths worldwide from World Health Organization (WHO) and increasing fast in low and middle income countries. Besides, these diseases pose an enormous negative impact on patients' quality of life with long-term disability and enormous health care cost. The prevalence and recurrence of chronic disorders are rising because of an aging population and high rates of poor lifestyle habits. In response to such development, the previous post event retro-diagnosis and passive medical mode are difficult to match the demands of modern public health, the establishment of risk in the early period and personalized intervention system for those high risky groups is the core direction of current medical research [1].

EHRs offer an unmatched data source for management of chronic diseases. A standard EHR system contains multimodal data that may be classified as structured test results, drug usage records, unstructured clinical notes, and medical images (X-rays, MRIs, CTs), and daily monitoring. These evidence are indicative of trajectory of the disease process and response to intervention of the patient and are important advocates in health risk modeling at the long term [2]. However, it is challenging for traditional unimodal or shallow machine learning techniques to obtain the effective global representation on the smpi data and construct the accurate model because of the vagaries of the illsuited data format, uneven time span, and the high level of noise and deletion. Thus, the way to construct a system that can deeply recognize and model the multimodal information has been a key breakthrough to enhance the performance that minimizes chronic disease risk prediction accuracy. Wang, Zhang, and Cheng [13] similarly highlight how AI-driven frameworks can reshape *real-time risk detection* in finance, showing the cross-domain importance of robust multimodal modeling.

In recent years, multimodal learning has risen as a vital segment of artificial intelligence research and has achieved great success in various tasks such as image-text retrieval, emotion recognition, and video understanding. Its growth in the area of medicine is also quickly progressing. Multimodal models have better fitness than traditional models which depend on a single modality, they are able to fully characterize the physiological and pathological changes, behaviors or lifestyles of patients based on the integrating various information of different modalities from multiple sources [3]. For instance, structured laboratory test data combined with unstructured physician diagnostic notes can promote the exchange of clinical semantics information and also help improve the model in the understanding of pathological progressive process. For the prediction of chronic diseases, multimodal learning also excels in capturing "soft signs" (potential pathology that has not yet fulfilled the disease criteria but has showed mild change), which lead to remarkable enhancement of the sensitivity and specificity in early screening to predict chronic diseases.

Large Language Models (LLMs) like GPT, BERT, T5 have achieved remarkable progress in the natural language processing domain, showing great proficiency for language understanding, generation and reasoning. Such models can comprehend the rich language and world information by pre-training in a large-scale and have some general task adaption ability. Nevertheless, existing traditional LLMs take only text as input and cannot perceive and model the non-verbal modalities, including images, time series, and physiological signals [4]. Given the fact that medical data are of tremendous diversity, this model restriction significantly limits the future application of LLMs into Medical DSSs. In a different setting, Feng, Dai, and Gao [14] proposed a reinforcement learning–enhanced LLM framework, showing how careful architectural design can greatly extend adaptability—an insight valuable for healthcare multimodal systems as well. As a result, it has paved the way to attempts to incorporate LLM into multimodal learning techniques to create a new model architecture that can handle heterogeneous data involving images and texts.

## II. Ease of Use

Li et al. [5] proposed a hierarchical model that adequately model long-range dependence for accurate prediction of chronic diseases, such as heart failure, diabetes, chronic kidney disease, and stroke. On the multimodal large-scale longitudinal EHR dataset, Hi-BEHRT achieved an improvement of 1-5% (AUROC) and 1-8% (AUPRC) over baselines. Additionally, Yang et al. [6] propose multimodal temporal-clinical note network to predict the mortality risk of ICU patients. The model integrates the time series data with clinical notes, and is able to guesstimating the evolution of patients' status by incorporating time perception mechanisms and attention mechanisms. Kline et al. [7] performed a systematic survey on multimodal machine learning in precision health. They studied 125 relevant articles and found that the multimodal approach is used particularly frequently in the fields of neurology and oncology. Lee et al. [8] proposed a multimodal deep learning model, including fundus images and conventional clinical risk factors (FRF), for the risk prediction of CVD. The model was evaluated on the SMC and UK Biobank data, and obtained AUROC of 0.781 and 0.872, respectively, demonstrating good modeling performance.

The HeLM (Health Large Language Model for Multimodal Understanding) model was introduced in Belyaeva et al. [9] proposes to utilize large language models (LLMs) for personalized health risk prediction. The model is capable of processing multiple data modalities through its modular design which supports the encoding of high-dimensional clinical data (e.g. time series data) into LLM-compatible embedding spaces. Li et al. [10] introduced a multimodal machine learning approach, which integrated electrocardiogram (ECG) and phonocardiogram (PCG) characteristics for cardiovascular disease forecasting. In this approach, a neural network model is built to get deep features of ECG and PCG respectively, the genetic algorithm for feature selection, and at last classified with support vector machine.

Agrawal et al [11] examined how to prune key predictors from 13,782 candidate multimodal features in an effort to improve the predictability of CAD events. The authors trained and tested 173,274 participants within UK Biobank, using the elastic web regularized Cox model, resulting in 51 predictors. Chaves et al. [12] constructed a multi-modality opportunistic risk assessment model for risk evaluation of IHD, in which such features were automatically extracted from abdominal CT images, and were used in conjunction with patient's electronic medical records (EMRs).

## III. Methodologies

### A. Cross-Modal Semantic Alignment

In VL-RiskFormer, we first collated the irregular follow-up records of patients with chronic diseases into a longitudinal sequence $\mathcal{E}_i = \{(x_{i,t}^{img}, x_{i,t}^{txt}, \Delta t_{i,t})\}_{t=1}^{T_i}$. Here, the $x^{img}$ is the chest x-ray, fundus, or wearable photograph, the $x^{txt}$ is the corresponding electronic medical record narrative, and the $\Delta t_{i,t} = t_{i,t} - t_{i,t-1}$ depicts adjacent visit intervals. The model is designed to learn to map $f_\Theta: \mathcal{E}_i \to (p_i(y), s_i, \hat{g}_i)$, where $p_i(y)$ gives the multi-disease risk distribution, $s_i$ is a confidence measure, $\hat{g}_i$ Individualized interventions are recommended. The whole training is done through joint losses, as Equation 1:

$$\mathcal{L}_{total} = \mathcal{L}_{CM} + \lambda_{time}\mathcal{L}_{TF} + \lambda_{graph}\mathcal{L}_{GAC} + \lambda_{sup}\mathcal{L}_{sup} + \lambda_{RL}\mathcal{L}_{RLHF}. \quad (1)$$

The above balances five types of signals: cross-modal alignment, time modeling, medical mapping, supervised fine-tuning, and enhanced feedback. We add momentum branches $\phi_{img}^m$ and $\phi_{txt}^m$ to the visual encoder $\phi_{img}$ and the text encoder $\phi_{txt}$, and update an exponential moving average, as Equation 2:

$$\phi^m \leftarrow \beta\phi^m + (1-\beta)\phi, \qquad \beta \in [0,1), \quad (2)$$

the larger $\beta$ causes the momentum branch to follow the main network slowly, ensuring that the negative sample cohort remains stable throughout the training period, thus providing a more reliable contrast background for rare lesions. In the cross-modal comparison loss, we use debiased InfoNCE to highlight the gradient contribution of difficult negative samples. For each positive pair $(z_k^{img}, z_k^{txt})$ and the negative sample $Q$ of the momentum library, the cosine similarity $sim(u,u) = u^\top v / \|u\| \|v\|$ and soft denominator $Z_k = \sum_{q \in Q} \exp(sim(z_k^{img}, q)/\tau)$ are calculated. Debias loss writing Equation 3:

$$\mathcal{L}_{CM} = -\frac{1}{B}\sum_{k=1}^{B} \log \frac{\exp\left(\frac{sim(z_k^{img}, z_k^{txt})}{\tau}\right)}{\exp\left(\frac{sim(z_k^{img}, z_k^{txt})}{\tau}\right) + \alpha Z_k}, \quad (3)$$

where $\alpha$ adaptive scaling according to the proportion of easily matched negative samples, so that the model focuses on difficult sample pairs with higher information content, so as to improve the resolution of early microscopic lesions. The cross-modal semantic alignment module projects images, texts, and time series into a unified embedding space through modal-specific encoders, and introduces a two-way hierarchical contrast loss function to ensure fine-grained semantic alignment between visual locals, key clinical phrases, and time segments. Similarly, Dai, Feng, Wang, and Gao [15] highlight the importance of ensemble-based strategies for multimodal customer identification, reinforcing the critical role of robust multimodal fusion in domains that require precise alignment. Irregular time intervals are embedded into the fusion feature by learnable position encoding, as in Equation 4:

$$\psi_{time}(\Delta t) = W_t \begin{bmatrix} \sin(\omega\Delta t), \cos(\omega\Delta t), \\ \Delta t, \log(1 + \Delta t) \end{bmatrix} + b_t. \quad (4)$$

This design captures both seasonal cycles (SinCosine), linearly decreasing effects $\Delta t$, and logarithmic compression to avoid long-interval information saturation, allowing the network to distinguish between "short-term rapid deterioration" and "long-term stable development". Subsequently, the visual and

textual encoded results are fused with time vectors in cross-attention modules, as in Equations 5 and 6:

$$z_t = CrossAttn\big(\phi_{img}(x_t^{img}), \phi_{txt}(x_t^{txt}), \psi_{time}(\Delta t_t)\big), \quad (5)$$

$$h_t = \mathcal{T}_{causal}(h_{<t}, z_t), \quad (6)$$

where causal Transformer $\mathcal{T}_{causal}$ employs a strict time mask and uses only historical information, avoiding training-deployment mismatch and maintaining clinical interpretability of disease course prediction. At the same time, we design a "hierarchical cross-attention mechanism" and a "modal adversarial network", the former realizes deep information fusion through the cross-attention of visual patches and text tokens, and the latter reduces the distribution difference between different modalities through the gradient inversion layer.

*B. Temporal Dynamics Modeling*

To explicitly inject medical prior, we compose the ICD-10 diagnostic code into a directed graph $G = (\mathcal{V}, \mathcal{E})$ and update the node embedding in *L*-layer graph attention, as Equations 7, 8, 9:

$$g_v^{(l+1)} = \sigma\left(\sum_{u \in \mathcal{N}(v)} \alpha_{vu}^{(l)} W_g^{(l)} g_u^{(l)}\right), \quad (7)$$

$$\alpha_{vu}^{(l)} = \frac{\exp(\gamma_{vu}^{(l)})}{\sum_{u'} \exp(\gamma_{vu'}^{(l)})}, \quad (8)$$

$$\gamma_{vu}^{(l)} = \big(g_v^{(l)}\big)^\top W_a^{(l)} g_u^{(l)}. \quad (9)$$

Based on ICD-10, we construct a disease map, with nodes coding diseases and edges representing known or learned co-occurrence relationships. The graph attention layer adopts a scaled dot product attention mechanism with relationship type embedding. Final aggregate representation $g^*$ is characterized by gated residuals injection sequences, as Equation 10:

$$\tilde{h}_t = h_t + \sigma(W_c[h_t \parallel g^*] + b_c) \odot g^*. \quad (10)$$

Enables the model to automatically consider comorbid chains such as diabetes, kidney disease, and heart failure when assessing risk, as Equations 11 and 12:

$$\hat{y}_c = \omega_c^\top \tilde{h}_{[CLS]} + b_c, \quad (11)$$

$$p(y = c) = \frac{\exp\left(\frac{\hat{y}_c}{T}\right)}{\sum_{c'}\left(\frac{\hat{y}_{c'}}{T}\right)}. \quad (12)$$

And in the inference stage, *M* dropout sampling is performed to calculate the variance, as shown in Equation 13:

$$s = Var_{m=1}^M p^{(m)}(y). \quad (13)$$

As a measure of uncertainty, high variance indicates the need for additional tests or manual review. In order to make the model output both accurate and feasible, we design a composite reward for the risk scoring network as the strategy $\pi_\theta$, as Equation 14:

$$R = \omega_1 AUROC + \omega_2(1 - ECE) + \omega_3 Act. \quad (14)$$

Use strategy gradients in combination, as Equation 15:

$$\nabla_\theta \mathcal{L}_{RLHF} = -\mathbb{E}_{\pi_\theta}\big[(R - \hat{b})\nabla_\theta \log \pi_\theta\big], \quad (15)$$

where $\hat{b}$ is the baseline. The model learns to automatically weigh between probabilistic calibration and clinical operability. In addition, temporal convolutional sublayers are introduced to capture more complex time dependencies, allowing the model to more accurately perceive the dynamic changes in condition.

IV. EXPERIMENTS

*A. Experimental Setup*

The experiment was based on the MIMIC-IV dataset, provided by BIDMC in the United States, covering structured, time-series, and textual electronic health records of more than 200,000 hospitalized and ICU patients. We selected patients aged ≥18 years with ICU records and extracted multimodal features such as demographic information, diagnostic codes, laboratory tests, physiological monitoring sequences, and free-text clinical notes. Figure 1 illustrates the visualization of the MIMIC-IV structured data, with patient age on the horizontal axis, length of stay (LOS) on the vertical axis, the size of the dots representing the number of chronic disease types diagnosed per patient, and color indicating gender.

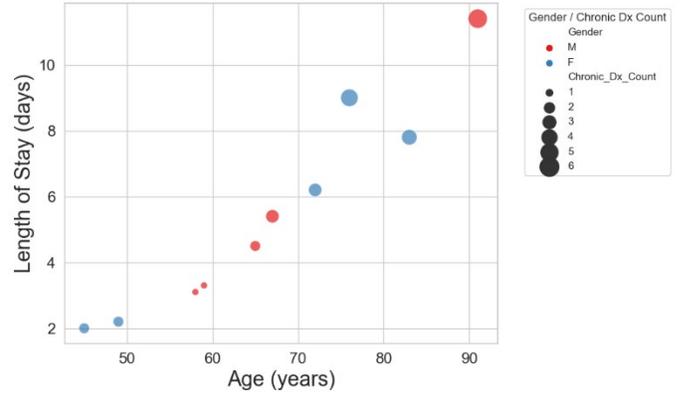

Fig. 1. Patient Age With the Length of Stay

We select four representative comparative approaches, covering different strategies ranging from structured electronic health records to multimodal data fusion including:

- Hi-BEHRT (Hierarchical Bidirectional Encoder Representations from Transformers) is a hierarchical Transformer model used for modeling longitudinally structured electronic medical record sequences.

- The MTNN (Multimodal Temporal-Note Network) consists of a two-branch structure: on one side is the LSTM network for modeling ICU time series.

- MM-ResNet (Multimodal Residual Network for CVD) is an image and structured feature fusion model for cardiovascular disease risk assessment. It uses ResNet50 to extract depth features of fundus images while introducing traditional risk scores as parallel inputs.

- MLP-MF (Multimodal Learning Pipeline with Medical Feature Selection) is a fully connected neural network architecture for multimodal feature selection and integration.

## B. Experimental Analysis

With the increase in the number of historical visits, the AUROC of all models showed an upward trend, indicating that more longitudinal visit information could help improve risk discrimination in Figure 2. Hi-BEHRT increased from about 0.66 to 0.77, MTNN from about 0.61 to 0.74, MM-ResNet and MLP-MF from about 0.60/0.58 to 0.70/0.67, respectively, and our VL-RiskFormer method continued to climb from about 0.68 to 0.84, always outperforming all other methods. After more than 60 visits in history, VL-RiskFormer has an advantage of 0.05–0.07 over sub-best model, highlights superior performance in deep integration of multimodal timing and domain priors. Comparable progress has been observed in other fields: Miao, Lu, and Wang [16] demonstrate that multimodal RAG frameworks can achieve significant improvements in housing damage assessment, highlighting the generalizability of multimodal architectures.

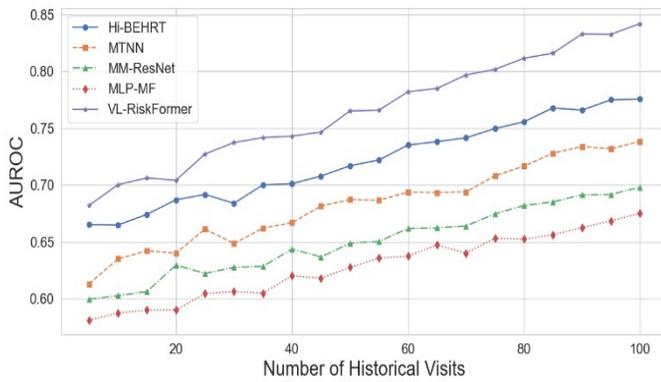

Fig. 2. Comparison of AUROC across Methods

As can be seen in Figure 3, there are significant differences in the individualized recommendations of LLMs for patients with different chronic diseases: patients with diabetes mainly receive "diet modification" and "exercise plan" recommendation, reflecting the high dependence of the disease on blood glucose management and lifestyle interventions. Hypertensive patients have the most concentrated suggestions on "stress management", which reflects the importance of psychological stress to blood pressure control. Patients with chronic kidney disease are more likely to receive "virtual follow-up" and "medication reminders" to ensure renal function monitoring and medication adherence.

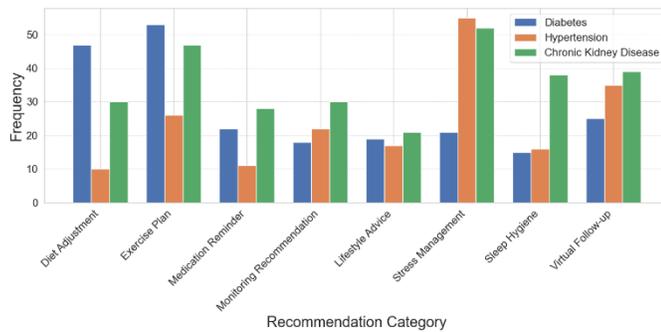

Fig. 3. VL-RiskFormer-Guided Recommendation Distribution by Disease Type

With the increase in the number of historical visits, the ECE of all models in Figure 4 showed a significant downward trend.

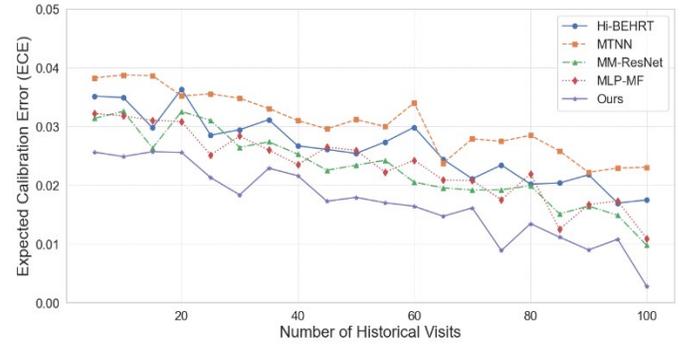

Fig. 4. Expected Calibration Error With Historical Visit

The ECE of Hi-BEHRT and MTNN decreased from about 3.5%–3.8% to 1.7%–2.3%, respectively, while that of MM-ResNet and MLP-MF decreased from about 3.2%–3.3% to 1.0%–1.1%. In contrast, the Ours method not only maintained the lowest calibration error across all visits, but also had a steeper descending curve, suggesting that it is more effective in utilizing multimodal historical records and RLHF calibration.

## V. CONCLUSION

In conclusion, the proposed VL-RiskFormer constructs an end-to-end multimodal chronic disease risk prediction and personalized intervention generation system by integrating structured clinical data, medical imaging, time series physiological signals and free text notes, combined with cross-modal contrast learning, time position coding, disease ontology map adaptation and RLHF optimization strategies. Future work can explore more efficient self-supervised cross-modal pre-training strategies to reduce dependence on labeled data. Equally important, Feng, Dai, and Gao [17] stress the need to account for personalized risks and regulatory safeguards in LLM deployment, an aspect highly relevant for future clinical applications.